\documentclass[pmlr,twocolumn,10pt]{jmlr} %

\usepackage{booktabs}
\usepackage{siunitx}
\usepackage{multirow}
\usepackage{enumitem}
\usepackage{breqn}
\usepackage{bbm}

\usepackage[switch]{lineno}

\theorembodyfont{\upshape}
\theoremheaderfont{\scshape}
\theorempostheader{:}
\theoremsep{\newline}

\jmlrvolume{297}
\jmlryear{2025}
\jmlrworkshop{Machine Learning for Health (ML4H) 2025} %

 \title[Let the Experts Speak]{Let the Experts Speak: Improving Survival Prediction \& Calibration via Mixture-of-Experts Heads}

\author{%
\Name{Todd Morrill} \Email{todd@cs.columbia.edu}\\
\addr Department of Computer Science, Columbia University, USA\\
\Name{Aahlad Puli} \Email{aahlad@nyu.edu}\\
\addr Department of Computer Science, New York University, USA\\
\Name{Murad Megjhani} \Email{mm5025@cumc.columbia.edu}\\
\addr Department of Neurology, Columbia University Medical Center, USA \\Department of Computer Science, Barnard College, USA\\
\Name{Soojin Park} \Email{sp3291@cumc.columbia.edu}\\
\addr Department of Neurology, Columbia University Medical Center, USA\\Department of Biomedical Informatics, Columbia University Medical Center, USA\\NewYork-Presbyterian Hospital at Columbia University Medical Center, USA\\
\Name{Richard Zemel} \Email{zemel@cs.columbia.edu}\\
\addr Department of Computer Science, Columbia University, USA
}

\begin{document}

\maketitle

\begin{abstract}
Deep mixture-of-experts models have attracted a lot of attention for survival analysis problems, particularly for their ability to cluster similar patients together. In practice, grouping often comes at the expense of key metrics such as calibration error and predictive accuracy. This is due to the restrictive inductive bias that mixture-of-experts imposes, that predictions for individual patients must look like predictions for the group they're assigned to. Might we be able to discover patient group structure, where it exists, while \emph{improving} calibration and predictive accuracy? In this work, we introduce several discrete-time deep mixture-of-experts (MoE)-based architectures for survival analysis problems, one of which achieves all desiderata: clustering, calibration, and predictive accuracy. We show that a key differentiator between this array of MoEs is how expressive their experts are. We find that more expressive experts that tailor predictions per patient outperform experts that rely on fixed group prototypes.
\end{abstract}

\paragraph*{Data and Code Availability}
In this work, we use three publicly available datasets. SUPPORT2 is a freely downloadable survival analysis dataset \citep{connors1995}.\footnote{\url{https://archive.ics.uci.edu/dataset/880/support2}} The Sepsis dataset \citep{reyna2020} is also freely available after a training course.\footnote{\url{https://physionet.org/content/challenge-2019/1.0.0/}} MNIST is a well-known research dataset that can be downloaded from a variety of sources.\footnote{\url{https://docs.pytorch.org/vision/main/generated/torchvision.datasets.MNIST.html}} Our GitHub code repository is available at \url{https://github.com/ToddMorrill/survival-moe}.

\paragraph*{Institutional Review Board (IRB)} Our research does not require IRB approval. 

\section{Introduction}
\label{sec:intro}
\begin{figure*}[ht!]
    \includegraphics{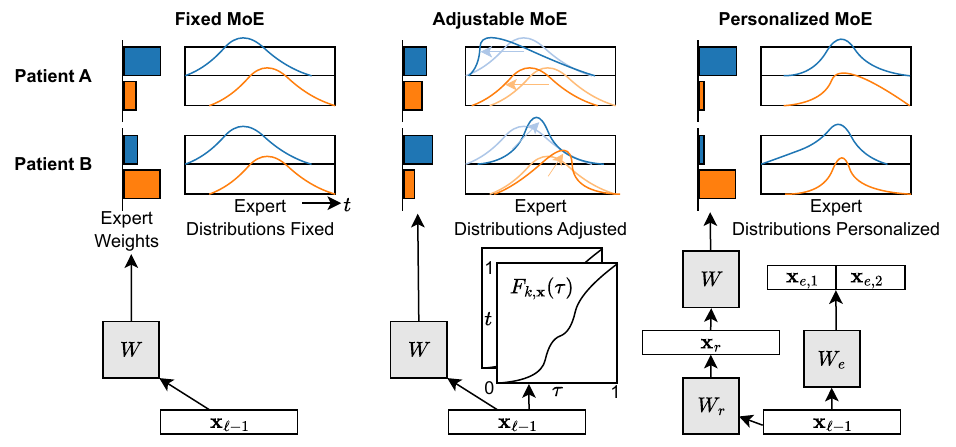}
    \caption{Illustration of our three proposed mixture-of-experts (MoE) architectures for survival analysis. Left - \textbf{Fixed MoE} showing the same expert distributions for patients A and B with differing expert weights. Middle - \textbf{Adjustable MoE} where fixed expert distributions (faint color) are adjusted per patient (dark color). Right - \textbf{Personalized MoE} where all experts produce a custom event distribution for all patients.}
    \label{fig:moe_illustration}
\end{figure*}

AI has the potential to have a profound impact on clinical decision support systems (CDSS). AI systems are being developed for medical imaging \citep{erickson2017}, disease diagnosis \citep{ahsan2022}, and many other clinical support settings. However, AI for CDSS faces barriers to adoption due to clinician mistrust of model predictions \citep{eltawil2023}. In this work, we focus on what clinicians care about most: highly accurate models, where probabilities have intuitive meaning (i.e., calibration), and interpretability of the model's decision-making process--in this case, the ability to reason by analogy to similar patients. Our work addresses survival analysis, where the task is to predict when clinical events will occur (i.e., time-to-event regression) while contending with right-censoring--not observing the event time for a subset of the patients.

Mixture-of-experts (MoE) models for medical survival analysis are particularly appealing for the above desiderata due to their ability to discover latent groups of patients. MoEs are defined by two key components: (i) a router that assigns patients to groups and (ii) a set of experts that produce event distributions for each group \citep{nagpal2021, hou2023, buginga2024}. Our goal is to investigate how expert expressivity in deep, discrete-time MoE heads affects calibration and predictive accuracy under matched capacity, which has not been carefully studied in the medical survival modeling setting. In our study, experts range from fixed prototypes to per-patient parameterizations. We therefore compare three MoE heads that differ only in expert expressivity to isolate its impact and include standard non-MoE baselines for context.

The first architecture, \textbf{Fixed MoE}, uses several experts where each expert learns an associated event distribution, which is then fixed across all patients. The second architecture, \textbf{Adjustable MoE}, again learns a prototypical event distribution per expert, but can be adjusted to the individual patient. The third architecture, \textbf{Personalized MoE}, uses experts that each form a custom event distribution for individual patients. Only the third architecture is able to cluster patients \emph{and} outperform strong baseline models on calibration, concordance index, and time-dependent Brier score. The contributions of our work are as follows: (i) we introduce three discrete-time deep MoE-based survival architectures, one of which achieves excellent clustering, calibration error, concordance index, and time-dependent Brier score and (ii) we report that the expressiveness of the experts is a key differentiator between discrete-time deep MoE-based survival analysis models, supported by a targeted set of experiments on these three architectures.

\section{Related Work}
\label{sec:related_work}

Building survival models with flexible model classes, such as Gaussian processes~\citep{alaa2017deep} or neural networks~\citep{ranganath2016deep}, and rich distribution families~\citep{miscouridou2018deep}, yields better predictions of risk than standard linear Cox or accelerated failure-time (AFT) models.
One can retain the structure of AFT models but improve prediction by using deep representations of the covariates~\citep{zhong2021,norman2024}.
Further generalizations that retain structure were introduced as conditional transformation models~\citep{hothorn2018,campanella2022}.
These choices improve expressivity to allow for better approximations of complicated survival distributions but do not reveal any structure in the data.

To better understand structure in the data, prior works ~\citep{manduchi2022,nagpal2021,nagpal2021a} cluster data, with each cluster assigned an ``expert'' distribution that is structured as Weibull-like or Cox-Proportional-Hazards-like curves.
\cite{buginga2024} instead work with unconditional Kaplan-Meier curves for each cluster assignment. However, clustering while restricting the expert distribution structure or using input-independent (unconditional) experts can hurt metrics like concordance~\citep{hou2023, buginga2024}.
While~\cite{manduchi2022,nagpal2021,nagpal2021a} do report improved metrics and clustering, fitting their models involves running variational inference techniques which require more work than is standard for using default supervised learning pipelines.
We work with discrete-time categorical distributions to remove limitations from Cox-like structures and study varying degrees of conditioning of experts to build calibrated and accurate survival models, without introducing additional complications to standard training. While prior work has addressed discrete-time survival modeling \citep{yu2011,fotso2018,lee2018,kvamme2021}, we are aware of no work addressing discrete-time deep MoE models for survival analysis.

There are similarities between MoE methods and ensemble-based approaches, particularly random survival forests \citep{ishwaran2008}. Both frameworks rely on aggregating/ensembling predictions from multiple specialized components (experts or trees). Recalibration is another notable line of work that aims to improve calibration in survival models via post-hoc methods (e.g., conformal prediction) \citep{qi2024}.

\section{Methods}
\label{sec:methods}

We now describe our three proposed deep mixture-of-experts (MoE)-based survival architectures. All models are feedforward deep learning models that use information from a patient's record (e.g., demographic data, physiological data, etc.) to forecast when they are likely to have a clinical event. Raw patient records $\mathbf{x}_0^i \in \mathbb{R}^a$ for patients $i=1\ldots N$ and $a$ the number of raw features are composed of categorical indicators (e.g., gender, etc.), and standardized continuous features (e.g., heart rate), meaning that for continuous features we subtract the mean and divide by the standard deviation. We learn embeddings for categorical indicators. After embedding lookups are concatenated with continuous features, this results in a $d$-dimensional feature vector that is fed to the model. Let $\mathbf{x} = \mathbf{x}^i_{\ell-1} \in \mathbb{R}^h$ be shorthand for the $h$-dimensional penultimate hidden state representation of our feedforward model with $\ell$ layers for the $i^{\text{th}}$ patient. We will now describe the $\ell^{\text{th}}$ layer (i.e., the final layer), which is the MoE head in all architectures. All methods are trained using a discrete-time Multitask Logistic Regression (MTLR) style loss function \citep{yu2011,fotso2018} that predicts a monotone label sequence (``once event occurs, it stays on''). This loss function has been shown to be well-calibrated \citep{haider2020}. The loss functions are described in Appendix Section \ref{sec:loss_functions}.

\subsection{Fixed Mixture-of-Experts}
The Fixed MoE architecture is representative of a class of prior works that use learned, but fixed per-patient, event distributions \citep{hou2023}. The Fixed MoE is composed of a learnable router $W \in \mathbb{R}^{n\times h}$ where $n$ is the number of experts and a collection of learnable experts $M \in \mathbb{R}^{n \times m}$, where each row can be mapped to a distribution over the $m$ possible discrete event times. We produce a probability mass function (PMF) $\mathbf{p}$ over discrete event times as follows
\begin{align}
    \boldsymbol{\alpha}(\mathbf{x}) &= \text{softmax}(\mathbf{x} W^{\intercal} / \kappa) \label{eq:expert_dist}\\
    \mathbf{p} &= \boldsymbol{\alpha} M' = \sum_{j=1}^n\alpha_jM'_j \label{eq:expert_pmf}
\end{align}
where $\alpha_j$ is the weight on expert $j$, $M'_j$ is the $j^{\text{th}}$ row of the parameter matrix $M$ after it has been normalized to form a discrete event distribution, and $\kappa$ is a learnable temperature parameter that modulates the sharpness of expert selection. To our knowledge, prior discrete-time neural survival models (e.g., DeepHit \citep{lee2018}) do not use an explicit MoE head over a categorical time grid, and MoE-style survival models have largely been continuous-time parametric or nonparametric \citep{hou2023,nagpal2021}.

\subsection{Adjustable Mixture-of-Experts}
The adjustable MoE can be seen as an exemplar from a class of models that transform a prototypical event distribution per expert to form a custom event distribution per patient \citep{nagpal2021,campanella2022,manduchi2022}. The adjustable MoE learns an event distribution per expert and then \emph{warps} it per patient. The following equations describe in detail how we warp prototypical event distributions but the important point is that with relatively few additional parameters per expert, we can flexibly adjust the event distributions per patient.

Let $\mathbf{t}\in[0,1]^m$ denote the canonical grid over the $m$ discrete time bins, with $t_j=j/(m-1)$ for $j=0,\ldots,m-1$. Each expert $k$ maintains a prototype vector $M_k\in\mathbb{R}^m$ of unnormalized scores over event times. To tailor expert $k$ to patient $i$, we define a strictly monotone bijection between the expert's internal time $\boldsymbol{\tau}\in[0,1]^m$ and the canonical grid $\mathbf{t}$:
\[
\underbrace{\phi_{k,\mathbf{x}}:\boldsymbol{\tau}\to\mathbf{t}}_{\text{forward}},\qquad
\underbrace{\psi_{k,\mathbf{x}}:\mathbf{t}\to\boldsymbol{\tau}}_{\text{inverse}}=\phi_{k,\mathbf{x}}^{-1}.
\]
Concretely, we take the forward map to be a normalized mixture of $r=2$ logistic cumulative density functions (CDF),
\begin{gather}
F_{k,\mathbf{x}}(u) = \sum_{r=1}^{2} w_{k,r}(\mathbf{x})\,\sigma\big(a_{k,r}(\mathbf{x})\,[\,u-c_{k,r}(\mathbf{x})\,]\big),\label{eq:two-logistic}\\
\sigma(z) = \frac{1}{1+e^{-z}},
\end{gather}
Weights $w_{k,r}(\mathbf{x})>0$ with $\sum_r w_{k,r}(\mathbf{x})=1$ enforce a linear combination of the two logistic CDFs that results in a function $F_{k,\mathbf{x}}(u)$ with a maximum value of 1. Slopes $a_{k,r}(\mathbf{x})>0$ control the steepness of each logistic component. Ordered centers $0<c_{k,1}(\mathbf{x})<c_{k,2}(\mathbf{x})<1$ control the location of each logistic component and we enforce an ordering on the centers so that the second logistic CDF is always to the right of the first. Importantly, $w_{k,r}(\mathbf{x})$, $a_{k,r}(\mathbf{x})$, and $c_{k,r}(\mathbf{x})$ are all learned linear functions of the final hidden state representation $\mathbf{x}$, allowing each patient to have a custom warping function per expert. We chose $r=2$ to allow for a more flexible warping of the event-time distribution compared to a single logistic CDF--see Figure \ref{fig:moe_illustration} for an example of the function's shape with $r=2$. Note that the left endpoint $F_{k,\mathbf{x}}(0)$ and right endpoint $F_{k,\mathbf{x}}(1)$ are not guaranteed to be 0 and 1, respectively, based on the parameterization in Equation \ref{eq:two-logistic}. We therefore enforce a mapping from $[0,1]\to[0,1]$ via endpoint normalization,
\begin{equation}
\tilde F_{k,\mathbf{x}}(u)\;=\;\frac{F_{k,\mathbf{x}}(u)-F_{k,\mathbf{x}}(0)}{F_{k,\mathbf{x}}(1)-F_{k,\mathbf{x}}(0)}\in[0,1],
\end{equation}
and define $\phi_{k,\mathbf{x}}(u)=\tilde F_{k,\mathbf{x}}(u)$ to be the forward map and $\psi_{k,\mathbf{x}}=\phi_{k,\mathbf{x}}^{-1}$ to be the inverse map. Given a canonical gridpoint $t_j\in [0, 1]$, we compute $u_{k,j}=(m-1)\,\psi_{k,\mathbf{x}}(t_j)$, which is an approximate index into the $m$ scores stored in $M_k$ on the model's internal time grid. We linearly interpolate the scores with
\begin{gather}
i_0 = \lfloor u_{k,j}\rfloor,\\
i_1=\min(i_0+1,m-1),\\
w_{k,j}=u_{k,j}-i_0,\\
\tilde M_{k,j} = (1-w_{k,j})\,M_{k,i_0} + w_{k,j}\,M_{k,i_1}.
\end{gather}
In practice, we evaluate $\psi_{k,\mathbf{x}}(t_j)$ via a bisection solver (see Appendix Section \ref{sec:inversion_and_gradients} for more details).

Let $\tilde M'_k$ denote the row-wise normalization of $\tilde M_k$ into an event-time PMF. The final per-patient PMF is then
\begin{align}
\boldsymbol{\alpha}(\mathbf{x}) &= \mathrm{softmax}\!\big(\mathbf{x}W^{\intercal}/\kappa\big),\\
\mathbf{p} &= \boldsymbol{\alpha} \tilde{M}' = \sum_{k=1}^{n}\alpha_k(\mathbf{x})\,\tilde M'_k.
\label{eq:adjustable_pmf_two_logistic}
\end{align}
This transformation family generalizes simple shift/scale warps and can emulate proportional-hazards style tilts while allowing richer early/late adjustments with minimal additional parameters per expert \citep{zhong2021,nagpal2021,campanella2022,manduchi2022}.

\subsection{Personalized Mixture-of-Experts}
The Personalized MoE architecture belongs to a collection of models that provide the most flexibility to the experts to form custom event distributions per patient (e.g., Deep Survival Machines \citep{nagpal2021a}). Since this architecture generates new expert distributions for each patient, we first project the final hidden state representation $\mathbf{x}$ to a router representation with $\mathbf{x}_r = \mathbf{x}W^\intercal_r$ for $W_r \in \mathbb{R}^{h\times h}$ and an expert representation with $\mathbf{x}_e = \mathbf{x}W^\intercal_e$ for $W_e \in \mathbb{R}^{h\times h}$. We then divide the expert representation into $n$ evenly-sized chunks $\mathbf{x}_{e,k}$ for $k=1,\ldots, n$, which are each fed to a linear layer denoted $L_k \in \mathbb{R}^{m \times (h/n)}$ to form an event-distribution to obtain $M_k(\mathbf{x}_{e,k}) = \mathbf{x}_{e,k}L_k^\intercal$, which collectively form the dynamic matrix of unnormalized densities over event times $M(\mathbf{x}_{e}) \in \mathbb{R}^{n\times m}$. The final PMF is then
\begin{align}
    \boldsymbol{\alpha}(\mathbf{x}_r) &= \text{softmax}(\mathbf{x}_r W^{\intercal} / \kappa)\\
    \mathbf{p} &= \boldsymbol{\alpha} M(\mathbf{x}_{e})' = \sum_{j=1}^n\alpha_jM(\mathbf{x}_{e})'_j \label{eq:personalized_pmf}
\end{align}

where $M(\mathbf{x}_{e})'_j$ is the $j^{\text{th}}$ row of the parameter matrix $M(\mathbf{x}_{e})$ after it has been normalized to form a discrete event distribution.

Our method is notable for its parameter efficiency due to chunking the expert representation and may force the model to use independent information for each expert, which may be beneficial for clustering and predictive accuracy.

\section{Experiments}
\label{sec:experiments}

\begin{figure}
    \centering
    \includegraphics{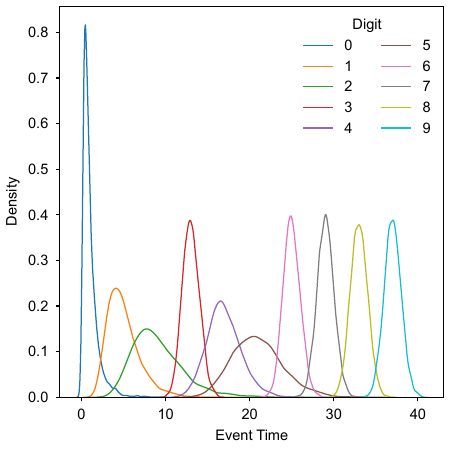}
    \caption{Event distributions for each digit in the Survival MNIST dataset. Each digit has a distinct event distribution, which allows us to evaluate the ability of our models to recover latent groups.}
    \label{fig:survival_mnist_event_distributions}
\end{figure}

We experiment with 3 datasets to probe the properties and capabilities of our proposed methods. Survival MNIST is a synthetic dataset (see Figure \ref{fig:survival_mnist_event_distributions}), which allows us to probe the models' abilities to predict event times and recover latent groups. We censor 15\% of examples. SUPPORT2 is a survival analysis dataset with 9,105 examples and $\sim$32\% censoring \citep{connors1995}. Sepsis is a larger dataset with 40,336 patient records and only 2,932 positive instances of sepsis, making this a very challenging anomaly detection survival analysis task \citep{reyna2020}. For the Sepsis dataset, the first 100 hours of each patient's ICU stay were summarized to perform a retrospective prediction, helpful when using partially labeled historical datasets to accelerate human labeling \citep{singer2016,henry2019}. Patients in the Sepsis dataset were administratively censored after 100 hours.

\begin{table*}
\caption{We report averages over 5 random seeds and in parentheses average deviation, given a random seed, from the MTLR baseline. Best results per dataset are \textbf{bolded}. $\downarrow$ indicates lower is better and $\uparrow$ indicates higher is better.}
\centering
\resizebox{\textwidth}{!}{
\begin{tabular}{llccccc}
\toprule
Dataset & Model & ECE $\downarrow$ & Concordance $\uparrow$ & Brier (25th) $\downarrow$ & Brier (50th) $\downarrow$ & Brier (75th) $\downarrow$ \\
\midrule
\multirow{6}{*}{Survival MNIST} & CoxPH & 0.030 (0.024) & 79.16 (-13.65) & 0.112 (0.083) & 0.159 (0.125) & 0.069 (0.059) \\
 & RSF & 0.057 (0.051) & 90.06 (-2.76) & 0.048 (0.019) & 0.073 (0.038) & 0.025 (0.015) \\
 & MTLR & 0.006 (0.000) & 92.82 (0.00) & \textbf{0.029 (0.000)} & \textbf{0.034 (0.000)} & \textbf{0.010 (0.000)} \\
 & Fixed MoE (ours) & 0.008 (0.002) & \textbf{93.46 (0.65)} & \textbf{0.029 (0.000)} & \textbf{0.034 (0.000)} & \textbf{0.010 (-0.001)} \\
 & Adjustable MoE (ours) & 0.008 (0.002) & 92.55 (-0.26) & 0.030 (0.001) & 0.037 (0.003) & 0.011 (0.001) \\
 & Personalized MoE (ours) & \textbf{0.005 (-0.001)} & 92.61 (-0.21) & \textbf{0.029 (0.000)} & 0.036 (0.002) & \textbf{0.010 (0.000)} \\
\cline{1-7}
\multirow{6}{*}{SUPPORT2} & CoxPH & 0.187 (0.130) & 78.89 (-1.03) & 0.212 (0.055) & 0.209 (0.060) & 0.236 (0.088) \\
 & RSF & 0.187 (0.129) & 79.76 (-0.15) & 0.207 (0.051) & 0.203 (0.055) & 0.232 (0.085) \\
 & MTLR & 0.057 (0.000) & 79.91 (0.00) & 0.156 (0.000) & 0.149 (0.000) & 0.148 (0.000) \\
 & Fixed MoE (ours) & 0.054 (-0.004) & 79.78 (-0.13) & 0.158 (0.001) & 0.147 (-0.002) & 0.145 (-0.003) \\
 & Adjustable MoE (ours) & \textbf{0.048 (-0.009)} & 79.83 (-0.08) & 0.158 (0.002) & 0.145 (-0.003) & 0.143 (-0.005) \\
 & Personalized MoE (ours) & \textbf{0.048 (-0.009)} & \textbf{80.84 (0.93)} & \textbf{0.154 (-0.002)} & \textbf{0.142 (-0.007)} & \textbf{0.138 (-0.009)} \\
\cline{1-7}
\multirow{6}{*}{Sepsis} & CoxPH & 0.635 (0.618) & 73.36 (-15.00) & 0.272 (0.253) & 0.541 (0.508) & 0.766 (0.727) \\
 & RSF & 0.604 (0.587) & 82.69 (-5.67) & 0.248 (0.230) & 0.603 (0.570) & 0.811 (0.773) \\
 & MTLR & 0.017 (0.000) & 88.36 (0.00) & 0.019 (0.000) & 0.033 (0.000) & 0.039 (0.000) \\
 & Fixed MoE (ours) & 0.011 (-0.006) & 87.09 (-1.27) & 0.019 (0.000) & 0.033 (-0.000) & 0.039 (0.001) \\
 & Adjustable MoE (ours) & 0.009 (-0.008) & 88.99 (0.63) & \textbf{0.017 (-0.001)} & 0.032 (-0.001) & 0.037 (-0.002) \\
 & Personalized MoE (ours) & \textbf{0.005 (-0.012)} & \textbf{89.77 (1.41)} & \textbf{0.017 (-0.002)} & \textbf{0.030 (-0.003)} & \textbf{0.036 (-0.003)} \\
\bottomrule
\end{tabular}
}
\label{table:results}
\end{table*}

\paragraph{Performance Assessment} As a comprehensive set of metrics covering performance with respect to accuracy and uncertainty calibration, we measure Harrell's concordance index \citep{harrell1996}, the time-dependent Brier score (standard inverse probability of censoring weighting (IPCW) \citep{gerds2006}) at 3 key points in time: the 25th, 50th, and 75th percentile of time bins, and equal mass expected calibration error (ECE) \citep{roelofs2022} adjusted with IPCW and averaged over all time bins. To control for parameter count as a potential source of performance differences, we ensure that all neural models have the same number of parameters (see Appendix Table \ref{tab:parameters}). All measurements are averaged over 5 random seeds. In order to obtain a ranking of the models by performance, \emph{for each random seed} we take the difference between the MoE model's metric and MTLR model's metric, and then average that over the 5 runs to obtain an average performance gap, which is reported in parentheses in Table \ref{table:results}. Network architectures and hyperparameters are described in Appendix Section \ref{sec:hyperparameters}. This experiment is important for two reasons: (i) it establishes calibration and predictive accuracy metrics for common non-clustering baseline models such as Cox proportional hazards (CoxPH) \citep{cox1972}, random survival forests (RSF) \citep{ishwaran2008}, and MTLR \citep{yu2011,fotso2018} against which any clustering-based method must be evaluated, and (ii) it allows us to compare performance across our three proposed MoE architectures, where the key difference between them is the expressivity of the MoE head.

\paragraph{Hyperparameter Sensitivity} The second set of experiments varies the number of experts in each MoE architecture to understand how the expressivity of the MoE head impacts the model's performance within each architecture. It also reveals any sensitivity to model specification (i.e., the number of latent patient groups). We vary the number of experts from 2 to 20 while holding all other hyperparameters constant.

\paragraph{Learned Patient Groups} We analyze the routing behavior of the MoE models to understand how patients are assigned to experts. Because Survival MNIST has well-defined latent groups, we can quantify the extent to which each expert specializes in a particular digit. Next, we assess the clinical relevance of patient clustering in a SUPPORT2 Personalized MoE model and quantify the stability of the routing behavior across random seeds using the Adjusted Rand Index (ARI) \citep{rand1971, hubert1985}.

\section{Results}
\label{Results}

\paragraph{Performance Assessment} We report our results in Table \ref{table:results}. The Survival MNIST dataset represents the Platonic ideal of a dataset with clear latent groups and as a result the Fixed MoE model performs best across all metrics except for calibration. This is a setting where the Fixed MoE is perfectly specified, with exactly 10 expert heads, one for each digit. The only information needed to make Bayes-optimal predictions is to identify the digit, at which point the appropriate expert can predict the group's event distribution. Any attempt to adjust or customize the expert distributions per patient only adds unnecessary complexity and can hurt performance. Nevertheless, the Personalized MoE model is best calibrated and matches the performance of the Fixed MoE on Brier at the 25th and 75th percentiles. The Survival MNIST dataset provides an interesting contrast to real-world datasets, where latent groups are almost never as well-defined.

Results on SUPPORT2 show the Personalized MoE model outperforms all other methods across all metrics. Importantly, the Personalized MoE model is outperforming the MTLR model on calibration error, concordance, and Brier score at all time points, while the fixed and adjustable MoE models are, in general, not. On SUPPORT2, our concordance and IPCW Brier scores are in line with prior reports for modern deep survival models (e.g., DSM \citep{nagpal2021a}). We see similarly excellent results on the Sepsis dataset, providing further evidence that the Personalized MoE model is able to deliver on all desiderata: clustering, calibration, and predictive accuracy. Consistent with prior work \citep{nagpal2021, manduchi2022}, we observe that per-patient adjustments can benefit calibration and predictive accuracy as measured by the Brier score relative to MTLR. Our Personalized MoE model extends those gains, delivering simultaneous improvements in calibration, accuracy, and discrimination, achieved with simple end-to-end learning rather than more complex training pipelines (e.g., spline estimation of baseline hazard rates \citep{nagpal2021}, EM-based cluster estimation \citep{buginga2024}, etc.). The Personalized MoE model's ability to form custom event distributions per patient appears to be key to its strong performance on real-world datasets, where latent groups are not as well-defined as in Survival MNIST.

\begin{figure}[htbp]
    \floatconts
    {fig:expert_sensitivity}
    {\caption{Expert sensitivity analysis over 5 random seeds varying the number of experts. Test set loss as a function of the number of experts.}}
    {%
        \subfigure[Survival MNIST]{\label{fig:expert_sensitivity_mnist}%
        \includegraphics{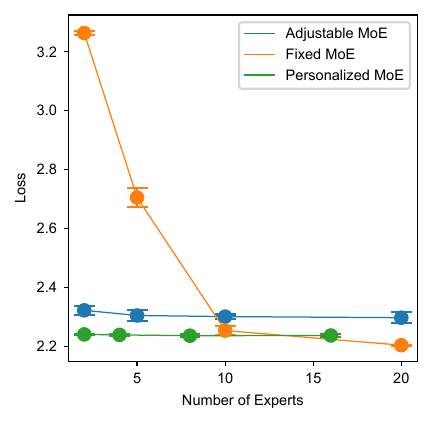}}%
        \qquad
        \subfigure[Sepsis]{\label{fig:expert_sensitivity_sepsis}%
        \includegraphics{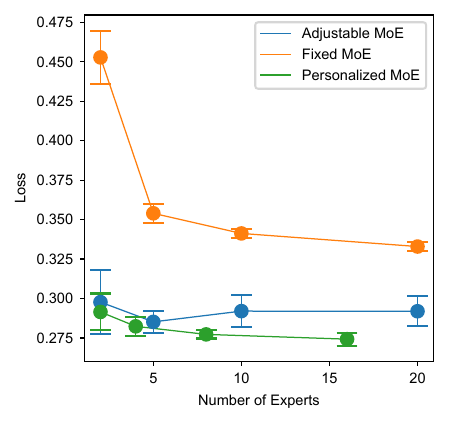}}
    }
    \label{fig:expert_sensitivity}
\end{figure}

\paragraph{Hyperparameter Sensitivity} We report the results of our expert sensitivity analysis in Figure \ref{fig:expert_sensitivity}. We observe that the Fixed MoE model is highly sensitive to the number of experts used before enough experts are present to capture the latent groups in the data. The adjustable MoE model is less sensitive to the number of experts, as it can adjust the expert distributions per patient even if there are insufficient number of experts. The Personalized MoE model is the least sensitive to the number of experts, as it can form custom event distributions for each patient regardless of the number of experts. We conclude that there is a continuum of sensitivity to model specification based on the expressivity of the experts.

\begin{figure}[ht!]
    \floatconts
    {fig:mnist_experts}
    {\caption{MNIST digit clustering by expert in (a) a \textbf{Fixed MoE} model and (b) a \textbf{Personalized MoE} model.}}
    {
        \subfigure[Fixed MoE Clustering]{\label{fig:fixed_mnist_experts}%
        \includegraphics{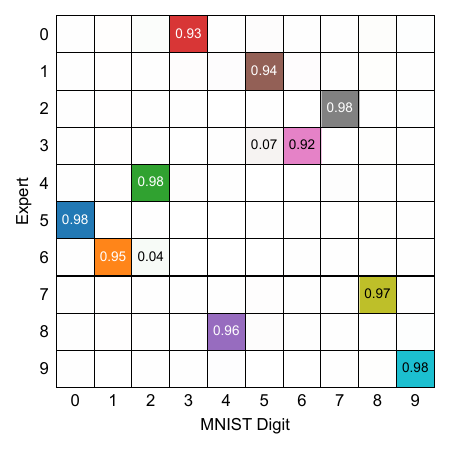}}%
        \qquad
        \subfigure[Personalized MoE Clustering]{\label{fig:personalized_mnist_experts}%
        \includegraphics{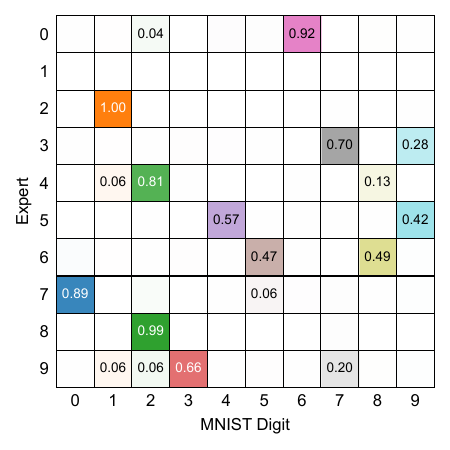}}
    }
    \label{fig:mnist_experts}
\end{figure}

\begin{figure*}
    \floatconts
    {fig:support2_expert_routing}
    {\caption{Routing of 911 unseen patients in the SUPPORT2 dataset to experts in a Personalized MoE model. (a) shows the group sizes (by Top-1 expert activation), (b) shows the survival curves for patients routed to each expert, and (c) shows the age distribution of patients in each cluster. The Personalized MoE model is able to discover clinically meaningful patient groups in real-world datasets by risk-profile and patient attributes.}}
    {%
        \subfigure[Group sizes]{\label{fig:expert_group_sizes}%
        \includegraphics[width=0.3\textwidth]{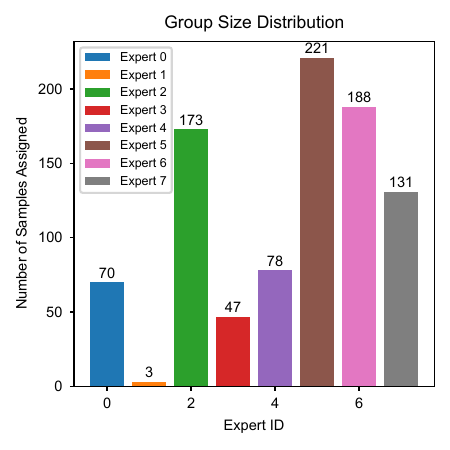}}%
        \qquad
        \subfigure[Survival curves]{\label{fig:expert_survival_curves}%
        \includegraphics[width=0.3\textwidth]{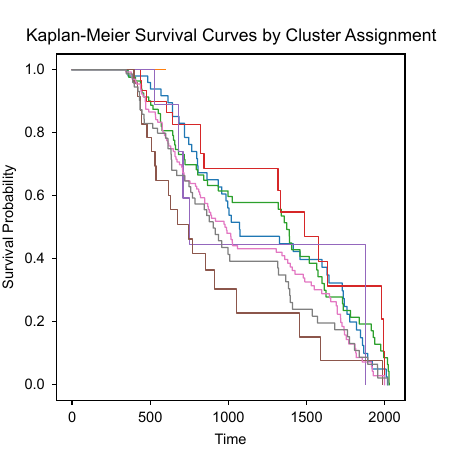}}
        \qquad
        \subfigure[Age by cluster]{\label{fig:expert_ages}%
        \includegraphics[width=0.3\textwidth]{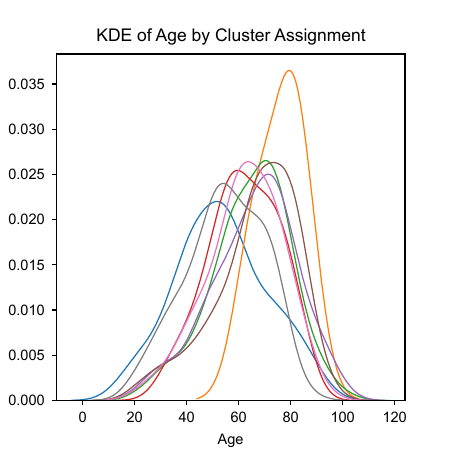}}
    }
    \label{fig:support2_expert_routing}
\end{figure*}

\paragraph{Learned Patient Groups} We report the first part of our routing analysis results in Figure \ref{fig:mnist_experts}, where each row of the matrices represents the distribution over Survival MNIST digits routed to that particular expert. For example, in Figure \ref{fig:fixed_mnist_experts}, among all data points that are routed to expert 0 in a Fixed MoE, 93\% are digit 3. Similarly, for expert 1, 94\% are digit 5, and so on. On the Survival MNIST dataset, we observe a strong degree of specialization per expert in the Fixed MoE model. The Personalized MoE model's routing decisions in Figure \ref{fig:personalized_mnist_experts} show a slightly lower, but still strong, degree of specialization per expert. For Figure \ref{fig:mnist_experts}, the routing decision for a data instance is made by selecting the expert with the highest weight. This degree of specialization per expert in both models indicates that these models are able to recover latent groups in the Survival MNIST dataset.

\begin{table}[h!]
\centering\small
\caption{Cluster sizes, Kaplan-Meier (KM) risk tier, and succinct signatures (clinician-informed). Vitals and labs are from day 3 of the hospital stay.}
\label{tab:cluster_signatures}
\begin{tabular}{p{1.9cm} p{1.1cm} p{4.0cm}}
\toprule
Cluster (Size) & KM risk tier & Signature \\
\midrule
$C_0$ ($n=70$)  & mid            & 50s; ARF/MOSF $\pm$ sepsis; low income (2nd lowest) \\
$C_1$ ($n=3$)   & n/a (tiny)     & 80s; metastatic lung cancer; anecdotal size \\
$C_2$ ($n=173$) & mid            & 70s; older White; lung CA/COPD/CHF/cirrhosis \\
$C_3$ ($n=47$)  & \textbf{lowest} & 50-60s men; metastatic colon CA; ESRD/HD with low ADLs; best vitals/pH \\
$C_4$ ($n=78$)  & varied         & 70s; older Black; ARF/MOSF $\pm$ sepsis; DNR (odd KM drop) \\
$C_5$ ($n=221$) & \textbf{highest}& 80s; dementia; \emph{early} DNR; ARF/MOSF $\pm$ sepsis/coma \\
$C_6$ ($n=188$) & mid            & 60s; COPD/CHF/cirrhosis; diabetes; \emph{lowest} income \\
$C_7$ ($n=131$) & elevated       & 50s; ARF/MOSF $\pm$ sepsis; coma/intubated; low income; 2-mo risk undercalled \\
\bottomrule
\end{tabular}
\end{table}

We also analyze the clustering behavior of a Personalized MoE model trained on SUPPORT2 in Figure \ref{fig:support2_expert_routing} to better understand how patient groups are formed on a real-world dataset. We see a variety of cluster sizes (Figure \ref{fig:expert_group_sizes}), indicating that Cluster 1 could possibly be merged into another cluster given its small size. We also find that patient risks are separated by cluster as evidenced by Figure \ref{fig:expert_survival_curves}. We then investigate kernel density estimates of continuous features (e.g., age in Figure \ref{fig:expert_ages}) broken down by cluster. For categorical features, we define a $z$-score based on the proportion of patients in a cluster that have a given categorical feature value relative to the overall proportion of patients with that feature value (see Appendix Section \ref{sec:cluster_zscores} for more details). Importantly, these analyses allowed us to define succinct, clinician-informed signatures for each cluster, which are summarized in Table \ref{tab:cluster_signatures} and described in more detail in Appendix Section \ref{sec:support2_clusters}. This indicates that the Personalized MoE model is able to discover clinically meaningful patient groups in real-world datasets.

In Appendix Figure \ref{fig:moe_predictions}, we display a detailed breakdown of two patients' routing behavior and predictions on all three MoE models, which provides some evidence that the routing behavior of the Fixed MoE and Adjustable MoE models is event time based, which is consistent with how these models are constructed.

We also measure the stability of the routing behavior for the Personalized MoE models across random seeds. We define stability as the degree to which patients are routed to the same expert across different random seeds. Using the Top-1 routing rule (the most active expert for the patient defines its cluster assignment), we find a moderate degree of stability as measured by the Adjusted Rand Index (ARI) with a value of 0.36, indicating that the Personalized MoE model is able to route patients to the same expert across different random seeds moderately consistently.

\section{Conclusion}
We introduce three discrete-time deep mixture-of-experts (MoE)-based survival architectures, one of which achieves excellent clustering, calibration error, and Brier scores at key points in time. We have shown that the expressiveness of the experts is a key differentiator between deep MoE-based survival analysis models, supported by a targeted set of experiments on these three architectures. We observe that the Personalized MoE model is able to deliver on all desiderata: clustering, calibration, and predictive accuracy as evidenced by results on three survival analysis datasets, including two real-world datasets. We explore sensitivity to the number of experts and find that the Personalized MoE model is the least sensitive to the number of experts, making it a robust choice for survival analysis tasks. We also analyze the routing behavior of the MoE models and use a synthetic survival analysis dataset to validate that the models are able to recover latent groups in the data. In practice, we find that the Personalized MoE model is able to discover clinically meaningful patient groups in real-world datasets. Overall, the Personalized MoE model provides a powerful and flexible approach to survival analysis that can adapt to the complexities of real-world clinical data and patient heterogeneity. It maintains strong performance across multiple metrics of interest to clinicians and researchers alike, including calibration and predictive accuracy, and enables reasoning by analogy to similar patients.

\section{Limitations}
This study does not claim state-of-the-art across all survival benchmarks or architectures. Our claims are relative and internal: within a controlled setting, increasing expert expressivity improves calibration and accuracy on two real-world datasets, with a synthetic counterexample (Survival MNIST) illustrating when fixed prototypes are optimal. Results should be interpreted in this scope; adding further model classes (e.g., DeepHit, continuous-time parametric mixtures, etc.) is orthogonal to our central question and left for future comparative work.

\acks{We would like to thank Jake Snell, Thomas Zollo, Zhun Deng, Kayla Schiffer-Kane, and Emily Saunders for their helpful discussions. This publication was supported by the National Center for Advancing Translational Sciences, National Institutes of Health, through Grant Number UL1TR001873. The content is solely the responsibility of the authors and does not necessarily represent the official views of the NIH. The authors also acknowledge support from the National Science Foundation and by DoD OUSD (R$\&$E) under Cooperative Agreement PHY-2229929 (The NSF AI Institute for Artificial and Natural Intelligence) (RZ). This work was also supported in part by National Institutes of Health: 1R01NS129760-01 (SP), 1R01NS131606-01 (SP), American Heart Association: 24SCEFIA1259295 (MM).}

\newpage
\bibliography{bib}

\clearpage
\appendix
\section{Additional Discussion \& Limitations}
There are limits to the number of experts that can be used in a Personalized MoE model for a given hidden dimension because: (i) the hidden dimension must be divisible by the number of experts, and (ii) the hidden dimension must be large enough to represent all discrete event times. Event times may cluster at key points in time, which would lower the dimensionality of the representation required to correctly decode. Our experiments work with 100 discrete event times and typically use a representation of size 16 (128 hidden dimensions / 8 experts), which is sufficient to represent the event time distribution for the datasets we explore, perhaps as a result of expert specialization by event time. If the event time distribution is more complex, then one may need a number of dimensions on the order of number of time bins to form a sufficiently expressive event distribution. Another limitation is that we do not explore time-varying inputs and outputs (i.e., multiple predictions per patient as their input features are updated). In principle, all of our methods can be attached to a recurrent neural network or Transformer but the details of how to interpret groups of patients in a time-varying setting is an open question. Another pro/con of the Personalized MoE is that it is not as sensitive to the number of experts as the other two models, which is limiting if the primary goal is to discover latent groups. One heuristic is to use the Fixed MoE to discover the number of latent groups (e.g., using the elbow method shown in Figure \ref{fig:expert_sensitivity}) and then use that number of experts in the Personalized MoE model. Another approach is to use a large number of experts in the Personalized MoE model and then prune/consolidate experts based on their routing behavior, but we leave this for future work. In future work, we are also interested in exploring simpler (yet expressive) transformation families for the Adjustable MoE that are monotone and naturally map from $[0,1]$ to $[0,1]$.

\section{Loss Functions}
\label{sec:loss_functions}
\subsection{Uncensored Loss}
All of our models are optimized using the Multitask Logistic Regression Loss (MTLR) \citep{yu2011,fotso2018}. We encode event times such that if patient $i$ has a medical event at time $s$ then all times $t\geq s$ will have label $y^i_{t}=1$. Correspondingly, times $t<s$ will be labeled with 0. Typical label strings will look like a sequence of 0s followed by a sequence of 1s (e.g., $(0, 0, 0, 1, 1)$). The probability that our model assigns to the ground truth label sequence for a particular patient (the likelihood function) is then given as
\begin{align}
p(\mathbf{Y}=(y_1, y_2, \ldots, y_m) \vert \mathbf{z}) = \frac{\exp(\sum_{j=1}^m y_j z_j)}{\sum_{k=0}^m \exp(f(\mathbf{z}, k))} \label{eq:mtlr_prob}
\end{align}
where $\mathbf{z}\in \mathbb{R}^m$ are logits from the model and $f(\mathbf{z}, k) = \sum_{j=0}^k 0\cdot z_j + \sum_{j=k+1}^m 1\cdot z_j$ for $0\leq k \leq m$, which constrains the space of valid event sequences to runs of 0s followed by runs of 1s and corresponds to the disease occurring in the interval $[k, k+1)$. The boundary case is $f(\mathbf{z}, m) = 0$, which corresponds to the sequence of all 0s. We can minimize the negative log-likelihood for an uncensored patient with the following loss
\begin{align}
    \mathcal{L}_{\text{uncensored}} =-\sum_{j=1}^m y_j z_j - \log \left(\sum_{k=0}^m \exp(f(\mathbf{z}, k))\right) \label{eq:mtlr_loss}
\end{align}
We now provide more detail on how to exchange between the model's logits $\mathbf{z}$ and a PMF $\mathbf{p} \in [0, 1]^m$ over event times. Our model produces \textit{increment logits} due to its interaction with the cumulative-sum parameterization of the softmax (Equation \ref{eq:mtlr_prob}). However, all methods must blend distributions over event times from different experts in probability space and therefore we rely on Equation \ref{eq:mtlr_prob} to map from increment logit space to probability space. We must also define an inverse operation to map from a PMF back to increment logits. We have
\begin{align}
    z_j = \log(p_j) - \log(p_{j+1}) \text{ for } j=1,\ldots, m-1 
\end{align}
and set $z_m = 0$. We now derive this inverse operation for completeness. Let 

\begin{align}
u_t = \sum_{j=t}^m z_j \text{ for } t=1,\ldots, m. \label{eq:logit_cumsum}
\end{align}

By Equation \ref{eq:mtlr_prob}, $p_j = \frac{\exp(u_j)}{\sum_{k=0}^m \exp(f(\mathbf{z}, k))}$ for $j=1,\ldots, m$, which is equivalent to $\text{softmax}(\mathbf{u})$ for $\mathbf{u} \in \mathbb{R}^{m}$. Recall that the softmax function is invariant to additive shifts (i.e., $\text{softmax}(\mathbf{u}) = \text{softmax}(\mathbf{u} + c)$ for any constant $c$). We can therefore set $c = -u_m$ so that $u_m = 0$ and by Equation \ref{eq:logit_cumsum}, $z_m = 0$. We can now write $u_j = \log(p_j) + c'$ for $j=1,\ldots, m$. Since $u_m = 0$, we have $c' = -\log(p_m)$ and therefore $u_j = \log(p_j) - \log(p_m)$ for $j=1,\ldots, m$. By Equation \ref{eq:logit_cumsum}, we have
\begin{align}
    z_j &= u_j - u_{j+1} \\
    &= \log(p_j) - \log(p_m) - \log(p_{j+1}) + \log(p_m) \\
    &= \log(p_j) - \log(p_{j+1}) \text{ for } j=1,\ldots, m-1
\end{align}
and $z_m = 0$, which completes the derivation.

\subsection{Censored Loss}
We now describe how to contend with censored data, which occurs when we do not observe if or when a patient has an event. Suppose a patient is censored at time $s_c$ and time $t + t_j$ is the closest time point after $s_c$. Then all sequences $\mathbf{Y} = (y_1, y_2, \ldots, y_m)$ with $y_i = 0$ for $i<j$ are consistent with this censored observation. Therefore the likelihood for a censored patient is the survival function (i.e., 1 minus the cumulative density function (CDF)), which is

\begin{align}
p(S \geq t+t_j \vert \mathbf{z}) = \frac{\sum_{k=j}^m\exp(f(\mathbf{z}, k))}{\sum_{k=0}^m \exp(f(\mathbf{z}, k))}    
\end{align}
and in turn, the negative log-likelihood loss for a censored patient is
\begin{dmath}
\mathcal{L}_{\text{censored}} =-\left[\log\left(\sum_{k=j}^m\exp(f(\mathbf{z}, k))\right) - \log \left(\sum_{k=0}^m \exp(f(\mathbf{z}, k))\right)\right].
\end{dmath}

\subsection{Regularizers}
We apply a load-balancing loss to all MoE models to ensure the model uses all available experts. Let $\bar{\boldsymbol{\alpha}} = \frac{1}{b}\sum_{i=1}^b \boldsymbol{\alpha}(\mathbf{x}^i)$ be the average expert distribution over a batch of size $b$. The load-balancing loss encourages the model to use all experts equally across the batch of examples by penalizing low-entropy average expert distributions. For a given batch, the loss is given as
\begin{align}
    \mathcal{L}_{\text{load-balance}} = \lambda_{\text{lb}} \cdot n\sum_{i=1}^n \bar{\alpha}_i^2,
\end{align}

where $n$ is the number of experts and $\lambda_{\text{lb}}$ is a hyperparameter that controls the strength of the regularization. This loss is minimized when $\bar{\boldsymbol{\alpha}}$ is the uniform distribution.

\section{Hyperparameters and Training Details}
\label{sec:hyperparameters}
\begin{table}[h]
\centering
\label{tab:parameters}
\caption{Parameter counts for all neural methods.}
\resizebox{\columnwidth}{!}{
\begin{tabular}{llc}
\toprule
Dataset & Model & Parameters \\
\midrule
\multirow{4}{*}{Survival MNIST} & MTLR & 187,189 \\
 & Fixed MoE & 209,844 \\
 & Adjustable MoE & 194,883 \\
 & Personalized MoE & 195,891 \\
\cline{1-3}
\multirow{4}{*}{SUPPORT2} & MTLR & 68,521 \\
 & Fixed MoE & 69,480 \\
 & Adjustable MoE & 69,435 \\
 & Personalized MoE & 62,141 \\
\cline{1-3}
\multirow{4}{*}{Sepsis} & MTLR & 62,945 \\
 & Fixed MoE & 63,008 \\
 & Adjustable MoE & 63,579 \\
 & Personalized MoE & 57,909 \\
\bottomrule
\end{tabular}
}
\end{table}

\begin{table}[h]
\centering
\caption{Model specific hyperparameters for Survival MNIST.}
\resizebox{\columnwidth}{!}{
\begin{tabular}{lp{0.75cm}p{1.3cm}p{1.6cm}p{1cm}}
\toprule
Model & Fixed\newline MoE & Adjustable MoE & Personalized MoE & MTLR \\
\midrule
Hidden Dim. ($h$) & 208 & 186 & 160 & 176\\
Num. Layers ($\ell$) & 2 & 2 & 1 & 2 \\
Num. Experts ($n$) & 10 & 10 & 10 & - \\
Learning Rate & 5e-4 & 5e-4 & 5e-4 & 5e-4 \\
\bottomrule
\end{tabular}
}
\end{table}

\begin{table}[h]
\centering
\caption{Model specific hyperparameters for SUPPORT2 dataset.}
\resizebox{\columnwidth}{!}{
\begin{tabular}{lp{0.75cm}p{1.3cm}p{1.6cm}p{1cm}}
\toprule
Model & Fixed MoE & Adjustable MoE & Personalized MoE & MTLR \\
\midrule
Hidden Dim. ($h$) & 176 & 186 & 128 & 176\\
Num. Layers ($\ell$) & 2 & 2 & 1 & 2 \\
Num. Experts ($n$) & 10 & 10 & 8 & - \\
Learning Rate & 5e-3 & 5e-3 & 5e-4 & 5e-4 \\
\bottomrule
\end{tabular}
}
\end{table}

\begin{table}[h]
\centering
\caption{Model specific hyperparameters for Sepsis dataset.}
\resizebox{\columnwidth}{!}{
\begin{tabular}{lp{0.75cm}p{1.3cm}p{1.6cm}p{1cm}}
\toprule
Model & Fixed MoE & Adjustable MoE & Personalized MoE & MTLR \\
\midrule
Hidden Dim. ($h$) & 176 & 186 & 128 & 176\\
Num. Layers ($\ell$) & 2 & 2 & 1 & 2 \\
Num. Experts ($n$) & 10 & 10 & 8 & - \\
Learning Rate & 5e-4 & 5e-4 & 5e-4 & 5e-4 \\
\bottomrule
\end{tabular}
}
\end{table}

\begin{table}[h]
\centering
\caption{Shared hyperparameters for all models.}
\begin{tabular}{lc}
\toprule
Hyperparameter & Value \\
\midrule
Batch Size & 64 \\
$\lambda_{\text{lb}}$ & 0.01 \\
$\kappa$ Init. & 2.0 \\
Time Bins ($m$) & 100 \\
\bottomrule
\end{tabular}
\end{table}

\begin{table}[h]
\centering
\caption{Random survival forest hyperparameters using the \texttt{scikit-survival} library.}
\resizebox{\columnwidth}{!}{
\begin{tabular}{lccc}
\toprule
Hyperparameter & Survival MNIST & Sepsis & SUPPORT2\\
\midrule
\texttt{n_estimators} & 100 & 50 & 200 \\
\texttt{max_features} & \texttt{sqrt} & \texttt{sqrt} & \texttt{sqrt} \\
\texttt{min_samples_split} & \texttt{100} & \texttt{50} & \texttt{200} \\
\bottomrule
\end{tabular}
}
\end{table}

\begin{table}[h]
\centering
\caption{Cox proportional hazards hyperparameters for all datasets using the \texttt{scikit-survival} library.}
\begin{tabular}{lc}
\toprule
Hyperparameter & Value \\
\midrule
\texttt{fit_baseline} & \texttt{True} \\
\texttt{alphas} & \texttt{[0.01]} \\
\texttt{l1_ratio} & \texttt{0.01} \\
\bottomrule
\end{tabular}
\end{table}
We define a validation set for Survival MNIST by randomly sampling 5,000 examples from the training set and then use the provided test set for final evaluation. For SUPPORT2 and Sepsis, we randomly sample 10\% of all examples to form a validation set and sample another 10\% to form a test set. We use the validation set loss to perform hyperparameter tuning for each dataset and method. The temperature parameter $\kappa$ is initialized to 2.0 for all MoE models and learned during training. The load balancing loss $\lambda_{\text{lb}}$ is constrained. For instance, setting the load balancing loss to be 0 or too close to 0 results in the model not using all of the experts. We found $\lambda_{\text{lb}}=0.01$ to consistently allow models to use most, if not all, the experts, while still allowing for specialization. We tuned the learning rate for all neural methods from the set $\{\text{5e-3}, \text{5e-4}, \text{5e-5}\}$. All models are trained with the Adam optimizer. For the Cox Proportional Hazards model we tuned the \texttt{alphas} hyperparameter from the set $\{0.001, 0.01, 0.1\}$ and the \texttt{l1_ratio} from the set $\{0.01, 0.5, 1.0\}$. The Random Survival Forest model's \texttt{n_estimators} hyperparameter was tuned from the set $\{50, 100, 200\}$ and the \texttt{min_samples_split} hyperparameter was tuned from the set $\{50, 100, 200\}$. Due to long runtimes and high memory requirements on the Survival MNIST dataset--likely due to the large number of features--we directly set \texttt{n_estimators} and \texttt{min_samples_split} to 100 and 100 respectively without tuning. We set hidden dimension sizes so that parameter counts are approximately equal across neural methods to ensure a fair comparison and isolate the effects of the decoder head used. We use a hidden dimension of 128-208 and 1-2 hidden layers for all models, which are fully connected layers followed by ReLU activations. We use 8-10 experts for all MoE models. The number of discrete time bins is set to $m=100$ for all datasets. We train all models to convergence as measured by the validation set loss, using early stopping with a patience of 10 epochs. All models are implemented in PyTorch and trained on a single GPU. Our GitHub code repository is available at \url{https://github.com/ToddMorrill/survival-moe}. We report the final hyperparameters used for each model and dataset in the tables above.

\section{Survival MNIST Data Generation}

\paragraph{Setup}
Let $\{(x_i,y_i)\}_{i=1}^N$ be MNIST images $x_i$ with class labels $y_i\in\{0,\dots,9\}$.
For each class $k$, we fix a target \emph{event-time} mean $m_k>0$ and standard deviation $s_k>0$ for a log-normal model of the event time $T\mid Y=k$.
In our experiments we use:

\begin{align*}
\boldsymbol{m}&=(1,5,9,13,17,21,25,29,33,37),\\
\boldsymbol{s}&=(1,2,3,1,2,3,1,1,1,1).
\end{align*}

\paragraph{Event-time model}
For class $k$, we draw a \emph{raw} event time
\begin{align*}
T^\star \,\big|\, (Y=k)\;\sim\; \mathrm{LogNormal}(\mu_k,\sigma_k^2),\quad\text{with pdf }\\
f_{T\mid k}(t)=\frac{1}{t\,\sigma_k\sqrt{2\pi}}\exp\!\left(-\frac{(\ln t-\mu_k)^2}{2\sigma_k^2}\right),\quad t>0,
\end{align*}
and CDF $F_{T\mid k}(t)=\Phi\!\left(\frac{\ln t-\mu_k}{\sigma_k}\right)$.
We choose $(\mu_k,\sigma_k)$ so that the resulting log-normal has \emph{mean} $m_k$ and \emph{standard deviation} $s_k$ on the original time scale. Using
$\mathbb{E}[T]=e^{\mu+\sigma^2/2}$ and $\mathrm{Var}(T)=(e^{\sigma^2}-1)e^{2\mu+\sigma^2}$, the code computes
\[
\mu_k=\log\!\left(\frac{m_k^2}{\sqrt{s_k^2+m_k^2}}\right),\qquad
\sigma_k=\sqrt{\log\!\left(1+\frac{s_k^2}{m_k^2}\right)}.
\]
(Equivalently, SciPy's \texttt{lognorm} is called with shape $s=\sigma_k$, scale $=\exp(\mu_k)$, and \texttt{loc}$=0$.)

\paragraph{Right-censoring}
We impose random right-censoring at rate $p_c=0.15$. Let $C\subset\{1,\dots,N\}$ be a uniformly random subset of size $\lfloor p_c N\rfloor$.
For $i\notin C$ (uncensored), we set the observed time $t_i=T_i^\star$ and event indicator $\delta_i=1$.
For $i\in C$ (censored), we draw a censoring time
\[
c_i \sim \mathrm{Uniform}(0,\,T_i^\star)
\]
and set $t_i=c_i$, $\delta_i=0$. Thus the observed triplet is $(t_i,\delta_i,y_i)$ with $t_i>0$ and $\delta_i\in\{0,1\}$.

\section{Evaluation Metrics}
\subsection{Equal-Mass Expected Calibration Error (ECE) for Discrete-Time Survival}

\paragraph{Notation}
We evaluate calibration over a discrete grid $\{t_1,\dots,t_T\}$ (here $T=100$).
For a batch of $N$ individuals, the model outputs \emph{increment logits} which are mapped to a discrete-time CDF via a monotone transformation:
\[
\widehat{F}_{ij} \;\equiv\; \widehat{F}(t_j\mid x_i)\in[0,1],
\]
where $\widehat{F}_{ij}$ follows from Equation \ref{eq:mtlr_prob}. The labels are $y_{ij}\in\{0,1\}$ with the \emph{monotone} convention (zeros up to the event time, then ones thereafter). Let $c_i\in\{0,1\}$ denote censoring ($c_i=1$ means censored) and $\delta_i=1-c_i$ the usual event indicator. Let $T_i$ denote the event time.

\paragraph{Equal-mass binning (per time $t_j$)}
Fix a number of calibration bins $Q$ (default $Q=10$). For each $t_j$:
\begin{enumerate}[leftmargin=1.5em]
\item Sort individuals by $\widehat{F}_{ij}$ in ascending order; denote the resulting permutation by $\sigma_j$.
\item Let $b=\lfloor N/Q\rfloor$ and $r=N\bmod Q$. Define bin sizes
$s_q=b+1$ for $q\in\{1,\dots,r\}$ and $s_q=b$ for $q\in\{r+1,\dots,Q\}$.
\item Assign the first $s_1$ sorted individuals to bin $q=1$, the next $s_2$ to $q=2$, etc. Define the one-hot assignment
$B_{ijq}=1$ if individual $i$ falls in bin $q$ at time $t_j$, else $0$.
Thus $n_{jq}\!=\!\sum_i B_{ijq}=s_q$ and $\sum_q n_{jq}=N$.
\end{enumerate}

\paragraph{Per-bin predicted probability}
Within each time $t_j$ and bin $q$,
\[
\overline{F}_{jq}
\;=\;
\frac{\sum_{i=1}^N B_{ijq}\,\widehat{F}_{ij}}{\sum_{i=1}^N B_{ijq}}
\;=\;
\frac{1}{n_{jq}} \sum_{i=1}^N B_{ijq}\,\widehat{F}_{ij}.
\]

\paragraph{Per-bin empirical event probability}
We estimate the empirical probability that the event has occurred by $t_j$ within bin $q$, denoted $\overline{Y}_{jq}$.

\paragraph{IPCW adjustment}
Let $G(\cdot)$ be the censoring survival, estimated by a Kaplan-Meier fit on the censoring process (using $c_i$ as the censoring ``event'' indicator). Kaplan-Meier estimates are implemented using \texttt{torchsurv} \citep{monod2024}. To avoid division by zero, evaluation times are clamped to the support of $G$.
Define event and survivor masks at each $(i,j)$:
\[
E_{ij}=\mathbbm{1}\{T_i\le t_j\}\,\delta_i,
\qquad
S_{ij}=\mathbbm{1}\{T_i> t_j\}.
\]
Use inverse-probability-of-censoring weights
\[
w^{\mathrm{evt}}_i=\frac{1}{G(\min\{T_i,\tau_G\})},\qquad
w^{\mathrm{surv}}_j=\frac{1}{G(t_j)},
\]
where $\tau_G$ is the largest time at which $G$ is defined. Then
\begin{align*}
\text{num}_{jq}&=\sum_{i=1}^N B_{ijq}\,w^{\mathrm{evt}}_i\,E_{ij},\\
\text{den}_{jq}&=\text{num}_{jq}+\sum_{i=1}^N B_{ijq}\,w^{\mathrm{surv}}_j\,S_{ij},
\end{align*}
and
\[
\overline{Y}_{jq} = \text{num}_{jq}/\text{den}_{jq},
\]

\paragraph{ECE aggregation (per time $t_j$)}
The equal-mass expected calibration error at time $t_j$ is the bin-count-weighted $\ell_1$ gap between predicted and empirical means:
\begin{align*}
\mathrm{ECE}(t_j) &= \sum_{q=1}^Q w_{jq}\,\bigl|\overline{F}_{jq}-\overline{Y}_{jq}\bigr|,\\
w_{jq} &= \frac{n_{jq}}{\sum_{q'=1}^Q n_{jq'}}=\frac{n_{jq}}{N}.
\end{align*}

(Weights are included because the last few bins may differ in size by at most one when $N$ is not divisible by $Q$.)

Finally, ECE can be averaged over all times $t_j$ to yield a single scalar summary statistic:
\[
\mathrm{ECE} = \frac{1}{T}\sum_{j=1}^T \mathrm{ECE}(t_j).
\]

\section{IPCW Brier Score for Discrete-Time Survival}

\paragraph{Notation} The setup and notation is similar to the ECE metric above.
\paragraph{Per-time squared residuals}
Define the per-entry squared residuals
\[
r_{ij} \;=\; \bigl(\widehat{F}_{ij} - y_{ij}\bigr)^2 .
\]

\paragraph{Inverse-probability-of-censoring (IPCW) weights.}
For evaluation at time $t_j$, use the standard IPCW weights
\[
w_i(t_j)
\;=\;
\frac{\mathbbm{1}\{T_i \le t_j\}\,\delta_i}{G(\min\{T_i,\tau_G\})}
\;+\;
\frac{\mathbbm{1}\{T_i > t_j\}}{G(t_j)}.
\]
In the implementation, arguments to $G$ are clamped to $\tau_G$ to avoid undefined values.

\paragraph{Brier score (IPCW)}
The IPCW Brier score at each time $t_j$ is
\[
\mathrm{BS}(t_j)
\;=\;
\frac{1}{N}\sum_{i=1}^N w_i(t_j)\, r_{ij}.
\]

Finally, the Brier score can be averaged over all times $t_j$ to yield a single scalar summary statistic:
\[
\mathrm{BS} = \frac{1}{T}\sum_{j=1}^T \mathrm{BS}(t_j).
\]

\subsection{Concordance Index}
We use the default settings for the concordance index (C-index) as implemented in the \texttt{SurvivalEval} library \citep{qi2023}. Their implementation uses uses linear interpolation of the discrete survival curve to find the median survival time for each patient. This value is then negated to form a risk score for each patient, which is then used to compute the C-index.

\section{IPCW-Adjusted Concordance Index}
Harrell's Concordance index (C-index) \citep{harrell1996} is widely used to evaluate the discriminative performance of survival models \citep{ranganath2016deep, khan2024}. At moderate levels of censoring (e.g., below 40\%) Harrell's C-index is a good estimator of the model's discriminative performance but at higher levels of censoring, it can be biased, which can be adjusted by using inverse probability of censoring weights (IPCW) \citep{polsterl2020}. In our study, SUPPORT2 has the highest rate of censoring at 32\%, so as a check on our results, we report IPCW-adjusted C-index in Table \ref{table:ipcw_cindex}.
\begin{table}[h]
\centering
\caption{IPCW-adjusted C-index on SUPPORT2. Parentheses show average deviation from MTLR baseline over 5 seeds.}
\resizebox{\columnwidth}{!}{
\begin{tabular}{lc}
\toprule
Model & IPCW-adjusted C-index $\uparrow$ \\
\midrule
CoxPH & 77.20 (-1.56) \\
RSF & 78.51 (-0.25) \\
MTLR & 78.75 (0.00) \\
Fixed MoE (ours) & 78.60 (-0.16) \\
Adjustable MoE (ours) & 78.27 (-0.49) \\
Personalized MoE (ours) & \textbf{79.67 (0.91)} \\
\bottomrule
\end{tabular}
}
\label{table:ipcw_cindex}
\end{table}

\section{Inversion and Gradients for the Two-Logistic Warp}
\label{sec:inversion_and_gradients}
\subsection*{A. Inversion by Bisection}

Recall the patient- and expert-specific forward map
\begin{align}
F_{k,\mathbf{x}}(u) &= \sum_{r=1}^{2} w_{k,r}(\mathbf{x})\,\sigma\!\big(a_{k,r}(\mathbf{x})[u-c_{k,r}(\mathbf{x})]\big),\\
\sigma(z) &= \tfrac{1}{1+e^{-z}},
\end{align}
and its endpoint-normalized version
\begin{equation}
\tilde F_{k,\mathbf{x}}(u)\;=\;\frac{F_{k,\mathbf{x}}(u)-F_{k,\mathbf{x}}(0)}{F_{k,\mathbf{x}}(1)-F_{k,\mathbf{x}}(0)}\in[0,1].
\end{equation}
We define
\[
\phi_{k,\mathbf{x}}(u)=\tilde F_{k,\mathbf{x}}(u)
\quad\text{and}\quad
\psi_{k,\mathbf{x}}=\phi_{k,\mathbf{x}}^{-1}.
\]
Since $\sigma$ is strictly increasing and $w_{k,r}(\mathbf{x}),a_{k,r}(\mathbf{x})>0$ with $0<c_{k,1}(\mathbf{x})<c_{k,2}(\mathbf{x})<1$, $F_{k,\mathbf{x}}$ (and hence $\tilde F_{k,\mathbf{x}}$) is strictly increasing on $[0,1]$, so the inverse exists and is unique.

For a given $(k,\mathbf{x},t_j)$ we obtain $\tau^\star=\psi_{k,\mathbf{x}}(t_j)$ as the unique solution to
\begin{align*}
    g(\tau;\theta)&=\tilde F_{k,\mathbf{x}}(\tau;\theta)-t_j\;=\;0,\\
    \theta&=(w_{k,1:2},a_{k,1:2},c_{k,1:2}).    
\end{align*}

Because $g(0)\le 0 \le g(1)$ and $g$ is strictly increasing, \emph{bisection} converges to $\tau^\star$ with bracketing on $[0,1]$.

\paragraph{Vectorized bisection (batched)} For all items in a batch and all experts/time-bins (indices suppressed):

\begin{enumerate}
\item Initialize $\text{lo}\leftarrow 0$, $\text{hi}\leftarrow 1$.
\item For $s=1,\ldots,S=20$:
\begin{enumerate}
\item $\text{mid}\leftarrow (\text{lo}+\text{hi})/2$.
\item $v\leftarrow \tilde F_{k,\mathbf{x}}(\text{mid})-t_j$.
\item Update $\text{lo}\leftarrow \mathbf{1}_{\{v<0\}}\!\cdot\text{mid} + \mathbf{1}_{\{v\ge 0\}}\!\cdot\text{lo}$, \ 
$\text{hi}\leftarrow \mathbf{1}_{\{v<0\}}\!\cdot\text{hi}  + \mathbf{1}_{\{v\ge 0\}}\!\cdot\text{mid}$.
\end{enumerate}
\item Return $\tau^\star \approx (\text{lo}+\text{hi})/2$.
\end{enumerate}

This procedure halves the bracketing interval at each step, so $S=20$ iterations yield $\approx 10^{-6}$ precision.

\paragraph{Numerical safeguards} We clip the denominator $D=F_{k,\mathbf{x}}(1)-F_{k,\mathbf{x}}(0)$ away from $0$ when forming $\tilde F_{k,\mathbf{x}}$, bound the slopes $a_{k,r}(\mathbf{x})\in[a_{\min},a_{\max}]$ (e.g., $a_{\min}=0.1$, $a_{\max}=35$), enforce $w_{k,1:2}$ via a softmax, and enforce ordered centers by a stick-breaking parameterization to improve conditioning.

\subsection*{B. Gradients via the Implicit Function Theorem}

Let $\tau^\star=\psi_{k,\mathbf{x}}(t_j)$ satisfy $g(\tau^\star;\theta)=0$ with $g(\tau;\theta)=\tilde F_{k,\mathbf{x}}(\tau;\theta)-t_j$. Because $\partial_\tau \tilde F_{k,\mathbf{x}}(\tau^\star;\theta)>0$, the implicit function theorem gives
\begin{equation}
\frac{\partial \tau^\star}{\partial \theta}
\;=\;
-\,\frac{\partial_\theta \tilde F_{k,\mathbf{x}}(\tau^\star;\theta)}{\partial_\tau \tilde F_{k,\mathbf{x}}(\tau^\star;\theta)}.
\label{eq:ift_core}
\end{equation}
Write $F=F_{k,\mathbf{x}}$, $F_0=F(0)$, $F_1=F(1)$, and $D=F_1-F_0$. Using
\[
\tilde F(\tau)=\frac{F(\tau)-F_0}{D},
\]
we obtain
\begin{align}
\partial_\tau \tilde F(\tau) &= \frac{\partial_\tau F(\tau)}{D},\\
\partial_\theta \tilde F(\tau) &= \frac{\partial_\theta F(\tau)\,-\,\partial_\theta F_0\,-\,\tilde F(\tau)\,\big(\partial_\theta F_1-\partial_\theta F_0\big)}{D}.
\label{eq:norm_grads}
\end{align}
For $F(\tau)=\sum_{r=1}^{2} w_r\,\sigma\!\big(a_r(\tau-c_r)\big)$ we have the elementary partials
\begin{align}
\partial_\tau F(\tau) &= \sum_{r=1}^{2} w_r\,a_r\,\sigma'\!\big(a_r(\tau-c_r)\big),\\
\sigma'(z)&=\sigma(z)\big(1-\sigma(z)\big), \\
\partial_{w_r} F(\tau) &= \sigma\!\big(a_r(\tau-c_r)\big),\\
\partial_{a_r} F(\tau) &= w_r\,(\tau-c_r)\,\sigma'\!\big(a_r(\tau-c_r)\big),\\
\partial_{c_r} F(\tau) &= -\,w_r\,a_r\,\sigma'\!\big(a_r(\tau-c_r)\big), &&
\end{align}
and the same forms evaluated at $\tau=0$ and $\tau=1$ for $F_0$ and $F_1$. Substituting \eqref{eq:norm_grads} into \eqref{eq:ift_core} yields closed-form expressions for $\partial \tau^\star/\partial \theta$.

\section{Discovered Patient Groups from a SUPPORT2 Personalized Mixture-of-Experts Model}
\label{sec:support2_clusters}

\paragraph{$C_0$ ($n=70$, mid KM)} Younger ARF/MOSF patients (often septic) from the second-lowest income stratum. KM initially tracks $C_4$ despite the age gap and then diverges later, consistent with later DNR decisions; physiologic severity is not extreme, so outcomes sit in the middle of the cohort.

\paragraph{$C_1$ ($n=3$, n/a (tiny))} Very elderly with metastatic lung cancer. The cell size is too small for inference, so we treat any apparent patterns as anecdotal.

\paragraph{$C_2$ ($n=173$, mid KM)} Older White patients with a cardiopulmonary/chronic-liver mix (lung cancer/COPD/CHF/cirrhosis). Outcomes are mid-range, worse than $C_3$ but better than the high-risk groups, consistent with substantial chronic disease burden without the acute derangements seen in $C_5$/$C_7$.

\paragraph{$C_3$ ($n=47$, best KM)} Middle-aged men with metastatic colon cancer and chronic disease (CHF/COPD/cirrhosis) but ESRD on hemodialysis may explain the creatinine right-tail without acute renal failure; ADLs are low. They show the best vitals (lower HR, good MAP), best pH and P/F ratio, least fever, and physicians also predicted short-term survival. Despite poor glucose control, this group has the best observed survival.

\paragraph{$C_4$ ($n=78$, varied KM)} Older Black patients with ARF/MOSF (often septic) and DNR status; the KM curve shows a bimodal/late drop that may reflect abrupt physiologic collapse uninterrupted due to DNR status. Clinicians labeled them ``risky in 2 months,'' and the early KM portion resembles $C_0$ (with less extreme outcomes) despite the age difference.

\paragraph{$C_5$ ($n=221$, worst KM)} Oldest cohort, dementia, coma, frequent malignancy/sepsis, and DNR very early after admission; HR is high, pH is acidotic, creatinine is markedly elevated.

\paragraph{$C_6$ ($n=188$, mid KM)} 60-something patients with COPD/CHF/cirrhosis and diabetes from the lowest income stratum. Despite comparatively favorable physiology in places and high predicted survival in other analyses, outcomes are worse than $C_3$; elevated HR.

\paragraph{$C_7$ ($n=131$, elevated KM)} 50-something patients with ARF/MOSF (± sepsis) who are often coma/intubated from the second-lowest income stratum; creatinine and fever are prominent, and HR is among the higher distributions. Mortality is high and clinicians undercalled 2-month risk relative to $C_4$, indicating under-recognized early hazard in this phenotype.

\section{Cell-wise $z$-scores for surfacing cluster-specific categories}
\label{sec:cluster_zscores}
For each categorical variable with levels $k\in\{1,\dots,K\}$ and cluster labels $c\in\{1,\dots,C\}$, let $n_{c,k}$ denote the count of observations in cluster $c$ having level $k$. Let row, column, and grand totals be
\[
n_c=\sum_{k} n_{c,k},\qquad
n_k=\sum_{c} n_{c,k},\qquad
N=\sum_{c,k} n_{c,k}.
\]
Under the independence (no-association) model, the expected count in cell $(c,k)$ is
\begin{equation}
e_{c,k} = \mathbb{E}[n_{c,k}] = \frac{n_c n_k}{N}.
\label{eq:expected}
\end{equation}
We compute the adjusted Pearson (Haberman) residuals
\begin{equation}
z_{c,k}
=
\frac{n_{c,k}-e_{c,k}}{\sqrt{e_{c,k}\bigl(1-\tfrac{n_c}{N}\bigr)\bigl(1-\tfrac{n_k}{N}\bigr)}},
\label{eq:haberman}
\end{equation}
which standardize observed-expected deviations.

\paragraph{Screening rule} We surface ``above-background'' cluster characteristics by flagging cells with $z_{c,k}>2$ (and, symmetrically, ``below-background'' with $z_{c,k}<-2$). In practice we report all cells with $\lvert z_{c,k}\rvert>2$.

\subsection{Flagged attributes by cluster}
\begin{figure*}
\centering
\includegraphics[width=\textwidth]{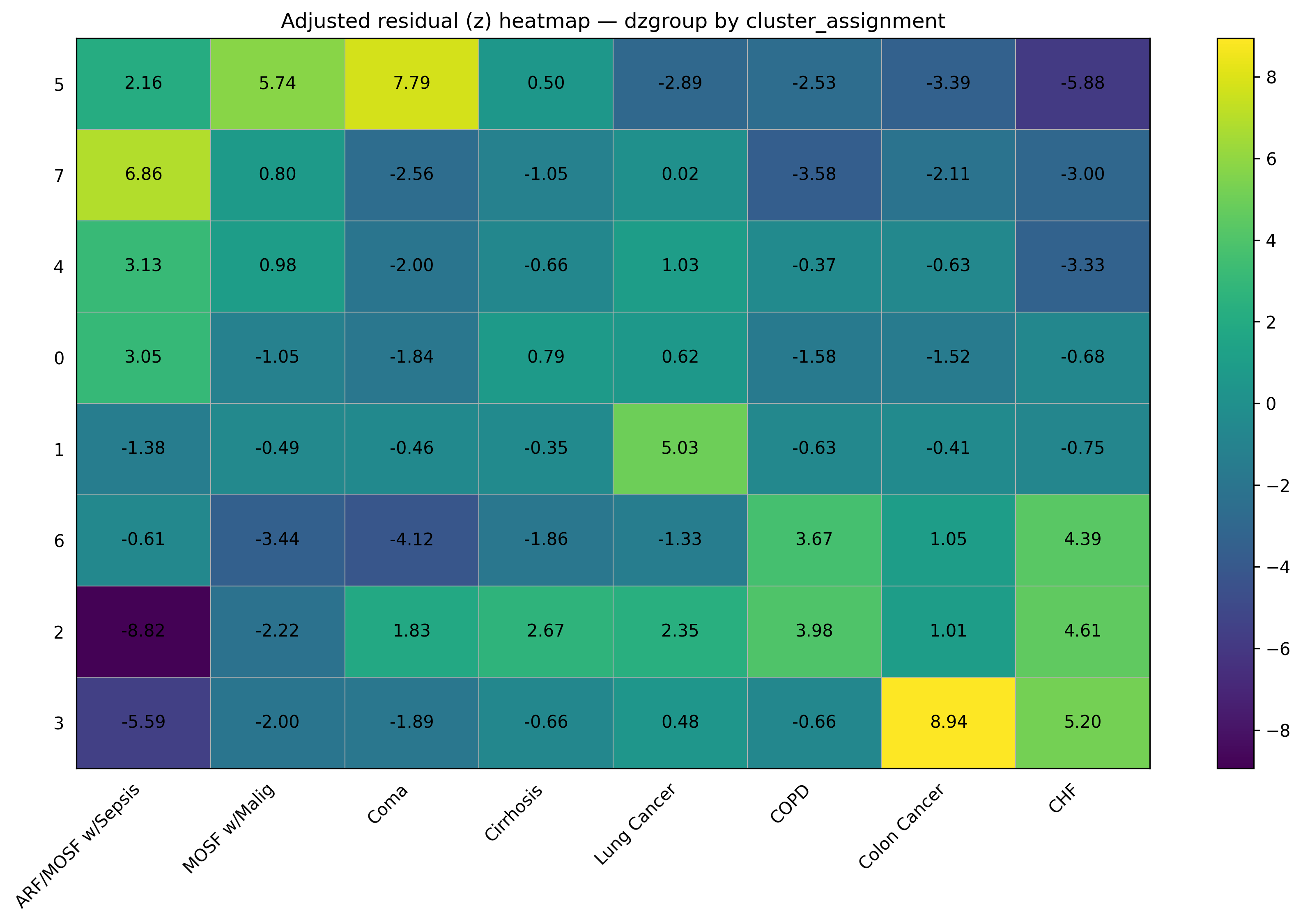}
\caption{Heatmap of $z$-scores for the \texttt{dzgroup} attribute in SUPPORT2 based on cluster assignments from the Personalized MoE model.}
\label{fig:support2_heatmap}
\end{figure*}

The following lists the flagged categorical variable levels for each cluster from the SUPPORT2 Personalized MoE model, along with their corresponding $z$-score values in parentheses. This analysis can also be performed visually using heatmaps of the $z$-scores as in Figure \ref{fig:support2_heatmap}.

\paragraph{Cluster 0 tags:}
\begin{itemize}
    \item death: 0 (value: 5.48)
    \item hospdead: 0 (value: 4.39)
    \item dzgroup: ARF/MOSF w/Sepsis (value: 3.05)
    \item dzclass: ARF/MOSF (value: 2.43)
    \item income: \$25-\$50k (value: 5.03)
    \item ca: no (value: 2.21)
    \item dnr: no dnr (value: 5.33)
    \item adlp: 1.0 (value: 2.27)
    \item sfdm2: no(M2 and SIP pres) (value: 3.22)
    \item sfdm2: missing (value: 4.15)
\end{itemize}

\paragraph{Cluster 1 tags:}
\begin{itemize}
    \item dzgroup: Lung Cancer (value: 5.03)
    \item dzclass: Cancer (value: 3.97)
    \item ca: metastatic (value: 3.39)
    \item adlp: 0.0 (value: 3.30)
    \item sfdm2: no(M2 and SIP pres) (value: 2.42)
\end{itemize}

\paragraph{Cluster 2 tags:}
\begin{itemize}
    \item hospdead: 0 (value: 6.30)
    \item dzgroup: Cirrhosis (value: 2.67)
    \item dzgroup: Lung Cancer (value: 2.35)
    \item dzgroup: COPD (value: 3.98)
    \item dzgroup: CHF (value: 4.61)
    \item dzclass: Cancer (value: 2.60)
    \item dzclass: COPD/CHF/Cirrhosis (value: 7.49)
    \item race: white (value: 2.87)
    \item diabetes: 0 (value: 2.45)
    \item dnr: no dnr (value: 4.32)
    \item adlp: 3.0 (value: 2.98)
    \item adlp: 2.0 (value: 2.12)
\end{itemize}

\paragraph{Cluster 3 tags:}
\begin{itemize}
    \item sex: male (value: 2.41)
    \item hospdead: 0 (value: 4.23)
    \item dzgroup: Colon Cancer (value: 8.94)
    \item dzgroup: CHF (value: 5.20)
    \item dzclass: Cancer (value: 5.91)
    \item dzclass: COPD/CHF/Cirrhosis (value: 3.35)
    \item ca: metastatic (value: 4.52)
    \item dnr: no dnr (value: 5.16)
    \item adlp: 2.0 (value: 4.25)
    \item sfdm2: no(M2 and SIP pres) (value: 4.13)
\end{itemize}

\paragraph{Cluster 4 tags:}
\begin{itemize}
    \item death: 1 (value: 5.07)
    \item hospdead: 1 (value: 6.24)
    \item dzgroup: ARF/MOSF w/Sepsis (value: 3.13)
    \item dzclass: ARF/MOSF (value: 3.57)
    \item race: black (value: 2.09)
    \item dnr: dnr after sadm (value: 12.72)
    \item adlp: missing (value: 2.35)
    \item sfdm2: $<$2 mo. follow-up (value: 6.99)
\end{itemize}

\paragraph{Cluster 5 tags:}
\begin{itemize}
    \item death: 1 (value: 8.59)
    \item hospdead: 1 (value: 18.60)
    \item dzgroup: ARF/MOSF w/Sepsis (value: 2.16)
    \item dzgroup: MOSF w/Malig (value: 5.74)
    \item dzgroup: Coma (value: 7.79)
    \item dzclass: ARF/MOSF (value: 5.13)
    \item dzclass: Coma (value: 7.79)
    \item income: missing (value: 2.00)
    \item dementia: 1 (value: 2.48)
    \item ca: yes (value: 4.68)
    \item dnr: dnr before sadm (value: 3.28)
    \item dnr: dnr after sadm (value: 15.98)
    \item adlp: missing (value: 11.33)
    \item sfdm2: $<$2 mo. follow-up (value: 16.70)
\end{itemize}

\paragraph{Cluster 6 tags:}
\begin{itemize}
    \item death: 0 (value: 6.80)
    \item hospdead: 0 (value: 9.07)
    \item dzgroup: COPD (value: 3.67)
    \item dzgroup: CHF (value: 4.39)
    \item dzclass: COPD/CHF/Cirrhosis (value: 5.19)
    \item income: \$11-\$25k (value: 3.10)
    \item diabetes: 1 (value: 3.55)
    \item dnr: no dnr (value: 10.58)
    \item adlp: 1.0 (value: 3.56)
    \item adlp: 0.0 (value: 12.00)
    \item sfdm2: no(M2 and SIP pres) (value: 8.86)
\end{itemize}

\paragraph{Cluster 7 tags:}
\begin{itemize}
    \item death: 0 (value: 2.43)
    \item hospdead: 0 (value: 4.02)
    \item dzgroup: ARF/MOSF w/Sepsis (value: 6.86)
    \item dzclass: ARF/MOSF (value: 7.12)
    \item income: \$25-\$50k (value: 2.29)
    \item dnr: no dnr (value: 5.29)
    \item dnr: missing (value: 3.45)
    \item sfdm2: SIP $>=$ 30 (value: 3.12)
    \item sfdm2: adl$>=$4 ($>=$5 if sur) (value: 5.23)
    \item sfdm2: Coma or Intub (value: 3.45)
\end{itemize}

\begin{figure*}[htbp]
\centering
\includegraphics[width=0.3\textwidth]{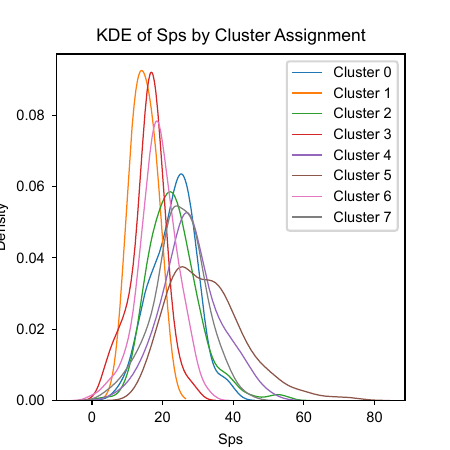}
\includegraphics[width=0.3\textwidth]{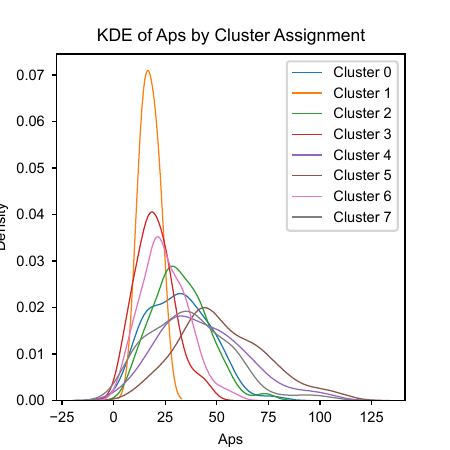}
\includegraphics[width=0.3\textwidth]{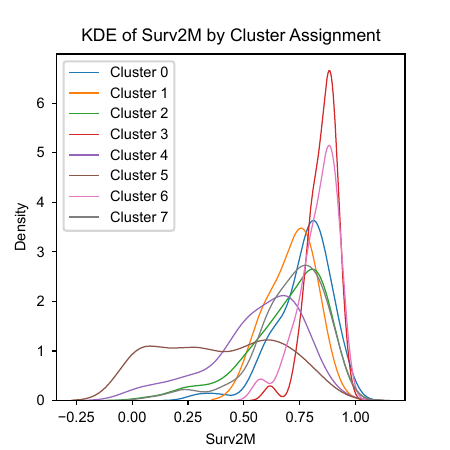}
\includegraphics[width=0.3\textwidth]{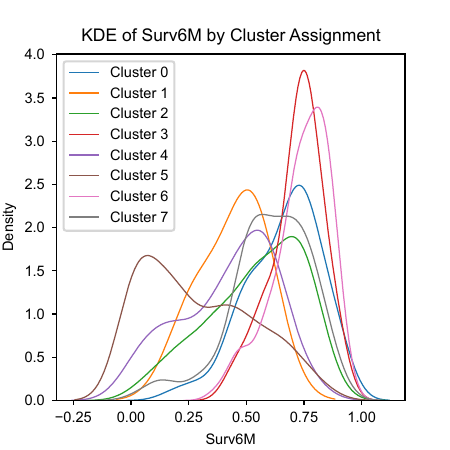}
\includegraphics[width=0.3\textwidth]{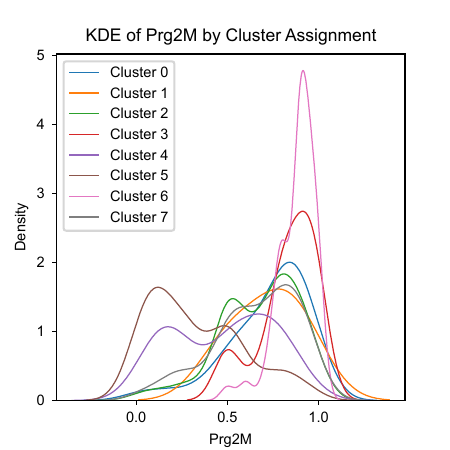} \includegraphics[width=0.3\textwidth]{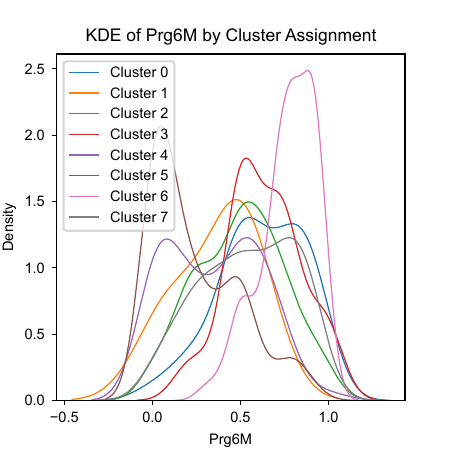} \includegraphics[width=0.3\textwidth]{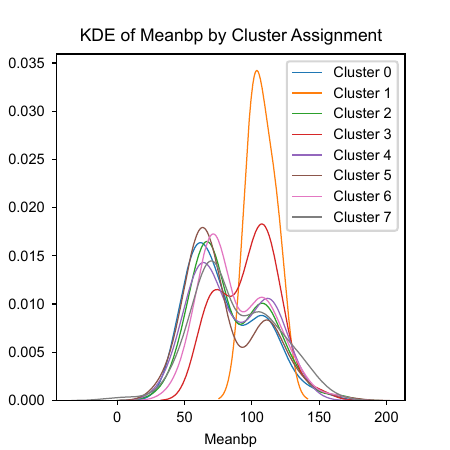}
\includegraphics[width=0.3\textwidth]{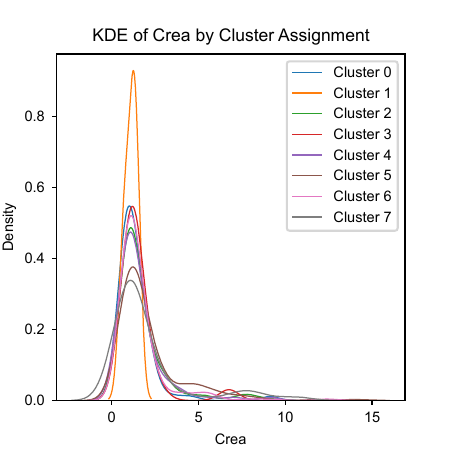} \includegraphics[width=0.3\textwidth]{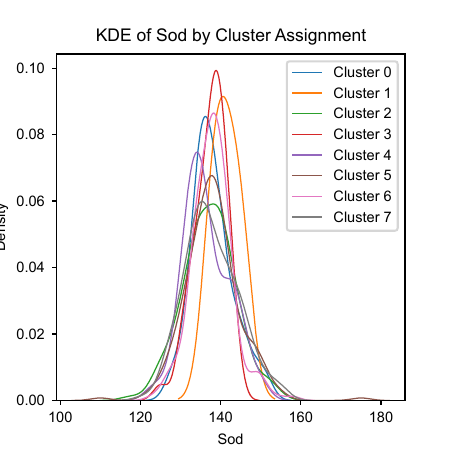} \includegraphics[width=0.3\textwidth]{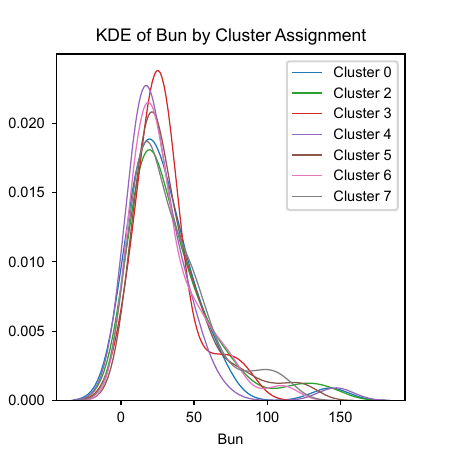}
\includegraphics[width=0.3\textwidth]{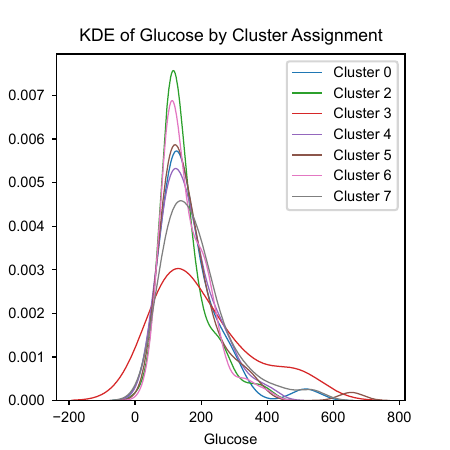} \includegraphics[width=0.3\textwidth]{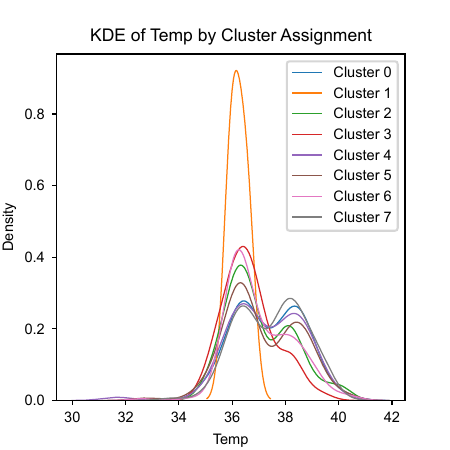} 
\caption{KDE plots for continuous variables by cluster assignment. Refer to data dictionary at \url{https://archive.ics.uci.edu/dataset/880/support2} for variable definitions.}
\end{figure*}

\section{Additional Figures}
\begin{figure*}[ht]
    \centering
    \includegraphics{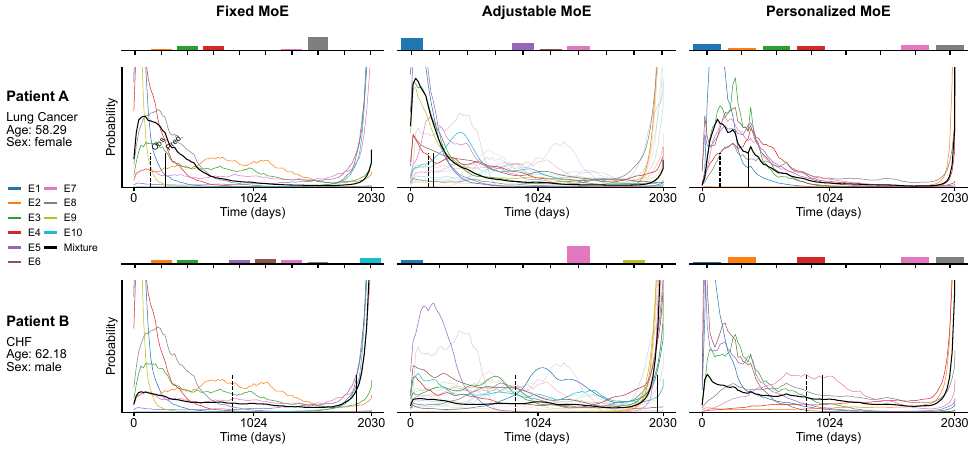}
    \caption{All plots from models trained on SUPPORT2. Left -  \textbf{Fixed MoE} model event distributions for each expert (faint colors) and adjustments (darker colors), middle - \textbf{Adjustable MoE} model event distributions for each expert, and right - \textbf{Personalized MoE} model event distributions for each expert.}
    \label{fig:moe_predictions}
\end{figure*}

\begin{figure*}[htbp]
    \centering
    \includegraphics{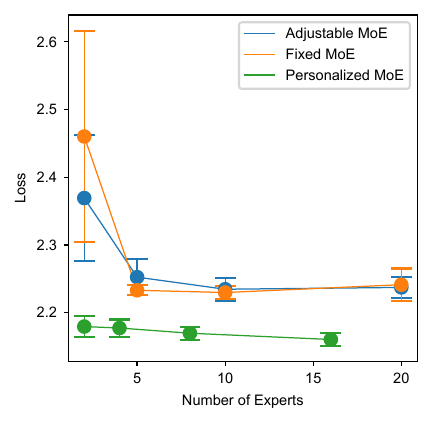}
    \label{fig:support2_expert_sensitivity}
    \caption{SUPPORT2 expert sensitivity analysis over 5 random seeds varying the number of experts. Test set loss as a function of the number of experts.}
\end{figure*}
\end{document}